# The Revisiting Problem in Simultaneous Localization and Mapping: A Survey on Visual Loop Closure Detection

Konstantinos A. Tsintotas, *Senior Member, IEEE*, Loukas Bampis, and Antonios Gasteratos, *Senior Member, IEEE*

*Abstract*—Where am I? This is one of the most critical questions that any intelligent system should answer to decide whether it navigates to a previously visited area. This problem has long been acknowledged for its challenging nature in simultaneous localization and mapping (SLAM), wherein the robot needs to correctly associate the incoming sensory data to the database allowing consistent map generation. The significant advances in computer vision achieved over the last 20 years, the increased computational power, and the growing demand for long-term exploration contributed to efficiently performing such a complex task with inexpensive perception sensors. In this article, visual loop closure detection, which formulates a solution based solely on appearance input data, is surveyed. We start by briefly introducing place recognition and SLAM concepts in robotics. Then, we describe a loop closure detection system's structure, covering an extensive collection of topics, including the feature extraction, the environment representation, the decision-making step, and the evaluation process. We conclude by discussing open and new research challenges, particularly concerning the robustness in dynamic environments, the computational complexity, and scalability in long-term operations. The article aims to serve as a tutorial and a position paper for newcomers to visual loop closure detection.

*Index Terms*—Loop closure detection, mapping, SLAM, visual-based navigation.

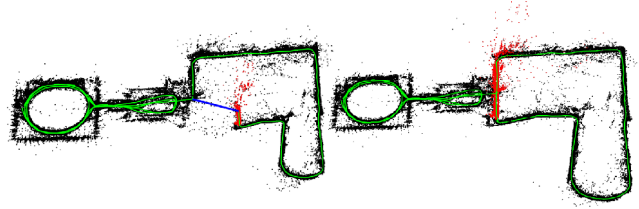

Fig. 1. A representative example of a pose graph constructed under a visual loop closure detection mechanism: map as generated initially (left) and subsequently a loop closure detection (right) in the New College dataset. [2]. The trajectory is drawn in green, while the chosen image pair is depicted in blue. When loops are detected, the system's accumulated drift error and uncertainty regarding the estimated position and orientation (pose) are bounded, allowing consistent map generation. Lastly, the robot's map is rectified extensively at both sides of the loop event (with permission from [3]).

## I. INTRODUCTION

LOOP closure detection, which has long been acknowledged as the primary rectification tool in any simultaneous localization and mapping (SLAM) system, historically represents a relevant and challenging task for the robotic community. Originally being introduced as "the revisiting problem," it concerns the robot's ability to recognize whether the sensory data just captured matches with any already collected, *i.e.*, a previously visited area, aiming for SLAM to revise its position [1]. As the accumulated dead-reckoning errors in the map may persistently grow when global positioning information is not available, loop closure detection is essential for autonomous navigation, mainly when operating in largely closed route scenarios. An important aspect is that loops inherently occur sparsely. Therefore, if the current sensor information does not match any previously visited location, the new observation is added to the robot's internal map, keeping only the constraints that relate the current with the penultimate pose. Since an erroneous loop closure detection might turn out to be fatal for any SLAM framework, a reliable pipeline should detect a small number or preferably zero false-positives while still avoiding false-negatives. The former refers to situations where the robot erroneously asserts a closed loop. The latter occurs when an event has been missed due to the system's misinterpretation. Hence, "closing the loop" is a decision-making problem of paramount importance for consistent map generation of unknown environments (Fig. 1).

Due to the above, the popularity of loop closure detection in the last 30 years is not surprising considering the notable SLAM evolution. Similar surveys can be found in the related literature, although they focus more on mapping techniques [4]–[6] or place recognition, relevant to robotics and other research areas, including computer vision, *viz.*, Lowry *et al.* [7], Zhang *et al.* [8], Masone and Caputo [9], and Garg *et al.* [10]. This article gives a broad overview of the loop closure pipelines in the last three decades, offering the perspective of how such a framework is structured into a SLAM system. We mainly go through a historical review of the problem, focusing on how it was addressed during the years reaching the current age.

### A. Foundation of Loop Closure Detection

In the early years, several kinds of methods were exploited to map a robot's environment, such as measuring bearings' revolutions and range finders; however, advances were limited by the computational resources and sensor capabilities available at the time. During the last two decades, researchers have been able to access to an enviable array of sensing devices, including massively produced multi-megapixel digital cameras and computers that are more potent in processing power and storage [11]. Images, which effectively capture the environment's appearance with high distinctiveness, are obtained through devices ranging from low-cost web cameras to high-end industrial ones. Not surprisingly, since modern robot navigation systems push towards effectiveness and efficiency, SLAM frameworks adopted such sensors and computational advances. Moreover, due to their reduced size and handiness, they can be easily attached to mobile platforms and allow the development of numerous localization and mapping pipelines with applications in different fields, such as autonomous cars [12], small aircrafts [13], and commercial devices [14].

Like any other computer vision task, visual loop closure detection firstly extracts distinct features from images; the similarities are then calculated, and finally, confidence metrics are determined. However, vital differences exist among image classification, image retrieval, and visual loop closure detection. More specifically, the first deals with categorizing a query image into a class from a finite number of available ones. Nevertheless, aiming to solve the nearest neighbor searching problem, image retrieval and visual loop closure detection systems face similar challenges as they try to find whether the current image matches any from the past. Due to this fact, an image retrieval module serves as the first part of any visual loop closure detection framework, in most cases. However, the underlying goals differ between these two areas regarding the sets upon which they operate. In particular, the latter searches for images depicting the exact same area that the robot is observing, operating only on the previously recorded image set. In contrast, image retrieval operates on an extensive database comprising not necessarily related images. This essentially means the typical goal is to retrieve instances of similar objects or entities, which may be different than the original in the query entry. For instance, a successful image retrieval could seek for buildings when the frame of a building is queried, or winter instances when the frame of a snowy road is used on a cross-season dataset. Hence, a considerable interest in the community's effort has been directed towards robust image processing techniques since sensory data representations, though appropriate for image retrieval, may not perform effectively in visual loop closure detection and vice versa.

Rather than working directly with image pixels, feature extraction techniques derive discriminative information from the recorded camera frames [15]. Hand-crafted descriptors, both global (based on the entire image) and local (based on a region-of-interest), were widely used as feature extractors. However, due to their invariant properties over viewpoint changes, local features were often selected for loop closure detection pipelines [16]. Deep learning has revolutionized many research areas [17], [18], with convolutional neural networks (CNNs) being used for various classification tasks as they can inherently learn high-level visual features. As expected, the robotics community exploited their capabilities in visual loop closure detection, especially in situations of extreme environmental changes [19]. Nevertheless, their extensive computational requirements limit their applicability in real-time applications and often induce the utilization of power-demanding general-purpose graphical processing units (GPGPUs) [20].

Through the extracted features, the robot's traversed path is described by a database of visual representations. To gain confidence about its position in the map and decide whether a loop occurs, the robot needs to compute a similarity score between the query and any previously seen observation. Several techniques exist for comparing images, ranging from pixel-wise comparisons to more complex ones based on feature correspondences. Then, a similarity threshold determines if a location can be considered as loop closure or should be declined, while additional steps, such as consistency checks based on multi-view geometry [21], can verify the matching pair. However, each of the aforementioned steps has to operate under real-time constraints during the robot's mission. With an increasing demand for autonomous systems in a broad spectrum of applications, *e.g.*, search and rescue [22], [23], space [24], [25] and underwater explorations [26], [27], the robots need to operate precisely for an extended period. As their complexity is at least linear to the traversed path, this limitation constitutes a crucial factor, severely affecting their capability to perform life-long missions.

In recent years, visual loop closure detection algorithms have matured enough to support continually enlarging operational environments. Thus, the research focus has shifted from recognizing scenes without notable appearance changes towards more complicated and more realistic changing situations. In such cases, detections need to be successful despite the variations in the images' content, *e.g.*, varying illumination (daytime against night) or seasonal conditions (winter against summer). Regardless of the advancements that have been achieved, the development of systems, which are condition invariant to such changes, remains an open research field. Finally, the growing interest of the robotics community is evidenced by the number of dedicated visual loop closure detection pipelines, as depicted in Fig. 2.

As we approach the third decade of visual loop closure detection, we need to acknowledge the groundwork laid out so far and build upon the following achievements:

1) Robust performance: visual loop closure detection can operate with a high recall rate in a broad set of environments and viewpoint changes (*i.e.,* different robot's orientations), especially when a location is revisited by a vehicle in the same direction as previously.

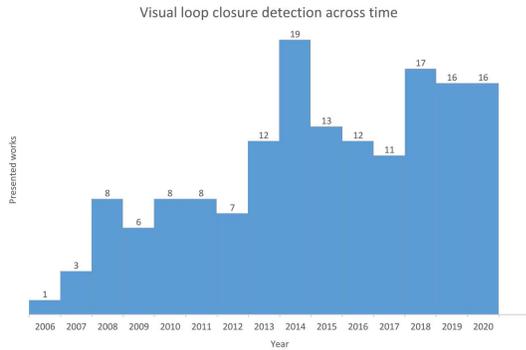

Fig. 2. An illustrative evolution histogram of loop closure detection pipelines, which is based on the methods cited in this article specifically addressing the visual loop closure detection task. Starting from the first approach [28] in 2006, the growth of appearance-based systems for indicating previously visited areas shows that they remain a growing research field. The peak observed in 2014 is highly related to the increased growth of visual place recognition in computer vision, along with the introduction of deep learning techniques.

2) **High-level understanding:** visual loop closure detection can extend beyond basic hand-crafted methods to get a high-level understanding and semantics of the viewing scene.
3) **Data management:** visual loop closure detection can choose useful perceptual information and filters out irrelevant sensor data to address different tasks. Moreover, it supports the creation of adaptive environment representations, whose complexity varies due to the task at hand.

### B. Paper Organization

Despite its unique traits, visual loop closure detection is a task inextricably related to visual place recognition. Thus, the article would not be complete unless it briefly examines the general concept of the latter in the robotics community (see Section II). Similarly, Section III provides a brief introduction to SLAM. The differences between the commonly used terms of localization, re-localization, and loop closure detection are also distinguished and discussed. In Section IV, an overview of the currently standard problem formulation for visual loop closure detection is given. The following sections review each of its modules in more detail. More specifically, Section V provides the methodology to describe the environment's appearance through feature extraction, including traditional hand-crafted and recent deep learning techniques. Section VI presents the environment's representation, *viz.,* database, and how locations are indexed during query. The robot's confidence originated from similarity metrics is addressed in Section VII, while Section VIII provides some benchmarking approaches. Section IX expands the discussion and examines the current open challenges in visual loop closure detection, *e.g.,* map scalability for long-term operations, recognition under environmental changes, and computational complexity. As a final note, considering this survey as a tutorial and positioning paper for experts and newcomers in the field, we prompt each reader to jump to the individual's section of interest directly. A map diagram of the topics discussed in the article at hand is depicted in Fig. 3. Light-grey boxes indicate the topics described in brief, while darker boxes are the ones presented in detail.

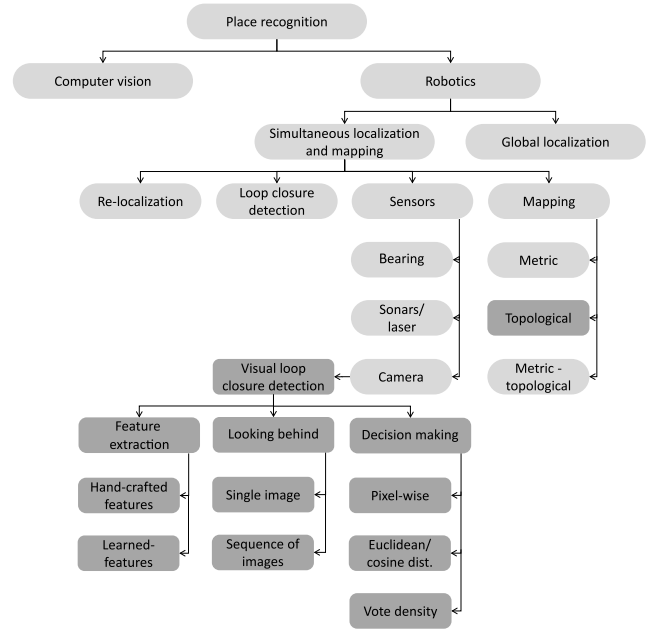

Fig. 3. Diagram depicting the taxonomy of topics discussed in the proposed article. Starting from the general concept of place recognition in computer vision and robotics, we describe how visual place recognition is used into SLAM pipelines to address the task of loop closure detection. The darker colored topics are the ones described in detail within this survey, while the lighter ones are briefly reported.

## II. VISUAL PLACE RECOGNITION

If we have been there before, we realize that viewing a single photograph is sufficient to understand where the picture was captured. This fact highlights the impact of appearance cues in localization tasks [29]–[32]. Historically, visual place recognition depicts a related task, studied intensively by the researchers in computer vision society within a broad spectrum of applications [33], including 3D reconstruction [34], map fusion [35], augmented reality [36], and structure-from-motion [37]. However, visual place recognition in robotics is somehow different. Since the knowledge of the environment, is a prerequisite for complex robotics tasks, it is vital for the vast majority of localization implementations or re-localization and loop closure detection pipelines within SLAM. Generally, it identifies the ability of a system to match a previously visited place using onboard computer vision tools. In robotics, the recognizer has to generalize as much as possible, in the sense that it should support robust associations among different recordings of the same place against viewpoint and environmental variations under run-time, storage, and processing power restrictions. Moreover, the application requirements for both domains are also driving this differentiation. The robotics community focuses on having highly confident estimates when predicting a revisited place, *viz.,* to perform visual loop-closure detection. At the same time, researchers in computer vision typically prefer to retrieve as many prospective matches of a query image as possible, *e.g.,* for 3D-model reconstruction [38]. More specifically, the former has to identify only one reference candidate associated with the previously visited traversal under varied conditions, while the latter can retrieve more matches, corresponding to a broad collection of images. Furthermore, in robotics, visual place recognition involves sequential imagery, significantly affecting the recognizer's performance. Finally, visual place recognition in robotics, apart from performing a topological constraint that indicates the same place in the database, produces geometric information that can be used to correct the trajectory, such as in the case of loop closure detection. In [7], Lowry *et al.* discuss the problem and provide a comprehensive survey of visual place recognition, while the work of Garg *et al.* [10] gives a broader discussion about the differences between visual place recognition in computer vision and robotics communities.

### III. SIMULTANEOUS LOCALIZATION AND MAPPING

A robot's capability to build a map (deriving the model of an unknown environment) and localizing (estimating its position) within that map is essential for intelligent autonomous operations and, during the last three decades, one of the most famous research topics [39]. This is the classic SLAM problem, which has evolved as a primary paradigm for providing a solution for autonomous systems' navigation without depending on absolute positioning measurements, such as the ones given by global navigation satellite systems (GNSS). Nevertheless, given the noise in the sensors' signal and modeling inaccuracies, drift is presented even if the most accurate state estimators are used. Therefore, the robot's motion estimation degenerates as the explored environment size grows, specifically with the traversed cycles' size therein [40]. A SLAM architecture commonly comprises a front-end and a back-end component. The former handles the unprocessed sensor data modeling that is amenable for estimation, and the latter performs assumptions based on the incoming sensory inputs. Loop closure detection belongs to the front-end, as it is required to create constraints among locations once the robot returns to an earlier visited area [3], while outlier (*i.e.,* false-positive loop closures) rejection is assigned to the back-end of SLAM [41]–[43]. In what follows, the role of loop closure detection and re-localization in the localization and mapping engines of SLAM is analyzed, and its dependence on the utilized sensing devices are examined.

#### A. Localization

Localization refers to the robot's task to establish its pose concerning a known frame of reference [44]. More specifically, "the wake-up robot problem," [45] *i.e.,* global localization, addresses the difficulty of recovering the robot's pose within a previously built map. At the same time, similar to the previous task, the re-localization task into SLAM, which is also known as the "kidnapped-robot problem," [46] concerns the position recovery based on a beforehand generated map following an arbitrary "blind" displacement, *viz.,* without awareness of the displacement, happening under heavy occlusions or tracking failures. The above tasks attempt to identify a correspondence connecting the robot's current observation with a stored one in the database. Nevertheless, the robot has no information about its previous pose during that process, and it can be considered lost. Contrariwise, constraints between the current and the previous pose are known during loop closure detection. More specifically, the system tries to determine whether or not the currently recorded location belongs to an earlier visited area and compute an additional constrain to further improve its localization and mapping (see Section III-B) accuracy. However, each of the aforementioned cases is addressed by similar mechanisms using the most recent observation and a place recognizer. If a match is successful, it provides correspondence and, in many cases, a transformation matrix between the current and the database poses in the map.

#### B. Mapping

Trajectory mapping, which is of particular interest in autonomous vehicles, provides the robot with a modeled structure to effectively localize, navigate, and interact with its surroundings. Three major mapping models exist within SLAM, *viz.,* metric, topological, and hybrid (metric-topological) maps. Metric maps provide geometrically accurate representations of the robot's surroundings, enabling centimeter-level accuracy for localization [47]. However, when the appearance information is not considered, more frequent loop closure detection failures in environments with repetitive geometrical structures are indicated. In addition, this model is also computationally infeasible when large distances are dealt with [48]. Relying on a higher representation level than metric ones, topological maps mimic the humans' and animals' internal maps [49]–[51]. A coarse, graph-like description of the environment is generated, where each new observation is added as a node, corresponding to a specific location. Furthermore, edges are used to denote neighboring connections, *i.e.,* if a location is accessible from a different one. This flexible model, introduced by Kuipers and Byun [52], provides a more compact structure that scales better with the traversed route's size. Regardless of the robot's estimated metric position, which becomes progressively less accurate, these approaches attempt to detect loops only upon the similarity between sensory measurements [53]–[55]. On the one hand, two nodes become directly connected during re-localization, enabling the robot to continue its mapping process. On the other hand, loop closure detection forms additional connections between the current and a corresponding node while the current node rectifying this way the accumulated mapping error [56] (see Fig. 4). An extensive review regarding topological mapping is provided by the authors in [5]. Finally, in metric-topological maps, the environment is represented via a graph-based model whose nodes are related to local metric maps, *i.e.,* a topological map is constructed, which is further split into a set of metric sub-maps [57]–[60].

#### C. Sensing

Aiming at overcoming GNSS limitations and detecting loop closures, different sensors have been used over the years,

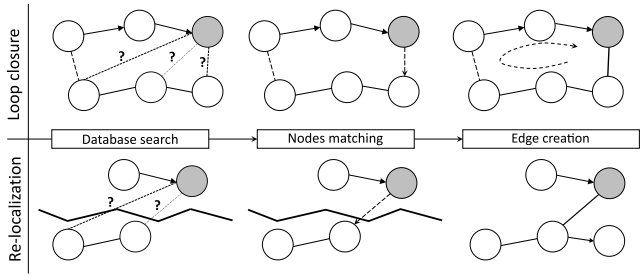

Fig. 4. A representative example highlighting the differences between topological loop closure detection and re-localization. The query node (shaded observation) searches the database for candidate matches and, subsequently, the most similar is chosen. Two components are connected through a constraint (bottom) when the system re-localizes its pose due to a tracking failure, while an another one edge is created between the two nodes (top) in the case of loop closure detection.

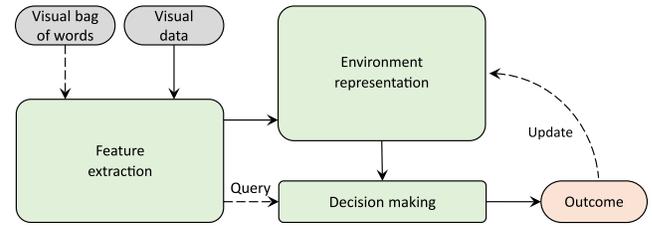

Fig. 5. Schematic structure depicting the essential parts of a visual loop closure detection system. The image is processed to extract the corresponding visual representation, by either using trained data (visual bag of words) or not, and the robot's internal database is constructed incrementally as the newly captured sensory measurement enters the system (visual data). When the query image arrives, its representation is compared against the database, *i.e.,* the environment representation, aiming to decide whether the robot navigates to an already visited area. Since loop closures occur sparsely, the database is updated accordingly when a match occurs.

including wheel encoders, sonars, lasers, and cameras. Generally, range finders are chosen because of their capability to measure the distance of the robot's surroundings with high precision [61]–[63]. However, they are also bounded with some limitations. The sonar is fast and inexpensive but frequently very crude, whereas a laser sensor is active and accurate; however, it is slow. Within the last years, since 3D maps [64] became more popular over traditional 2D [65], light detection and ranging (LiDAR) is established as the primary sensor for large-scale 3D geometric reconstructions [66]. Yet, they remain unsuitable for mass installation on mobile robots due to their weight, price, and power consumption. Furthermore, its measurements, *i.e.,* scan readings, cannot be distinguished during loop closure detection from locations with similar shapes but different appearances, such as corridors. Although successful mapping techniques based on range-finders are implemented [67]–[69], these types of sensors tend to be associated with, or replaced by, single cameras [70]–[76] or stereo camera rigs [77]–[82]. This is mainly due to the rich textural information embedded in images, the cameras' low cost, and their applicability to various mobile robots with limited computational powers, such as the unmanned aerial vehicles (UAVs) [83]. Yet, even if multi-sensor frameworks [84]–[87] can improve performance, especially in changing environmental conditions [88], such a setup requires expensive hardware and additional calibration than camera-only ones [89]. During the second decade of visual-based navigation, autonomous robots' trajectory mapping of up to 1000 km has been successfully achieved using cameras as the primary sensory modality [90]. The solution of a monocular camera provides practical advantages concerning size, power, and cost, but also several challenges, such as the unobservability of the scale or state initialization. Nevertheless, these issues could be addressed by adopting more complex setups, such as stereo or RGB-D cameras [91]–[94]. Lastly, even if limited attention has been received regarding the event-based cameras within the vision research community [95], their high dynamic range and the lack of motion blur are proved beneficial in challenging lighting conditions and high-speed applications [96].

## IV. STRUCTURE OF A VISUAL LOOP CLOSURE DETECTION SYSTEM

A loop closure detection system's generic block diagram is depicted in Fig. 5. Firstly, a system interpreting the environment's appearance has to detect previously visited locations by employing only visual sensory information; thus, the perceived images have to be interpreted robustly, aiming for an informatively built map. Then, the system's internal environment representation of the navigated path needs to be addressed. In many cases, such representations are driven by the robot's assigned mission. Aiming to decide whether or not the robot navigates a previously seen area, the decision extraction module performs data comparisons among the query and the database instances. Confidence is determined via their similarity scores. Lastly, as the system operates on-line, the map is updated accordingly throughout the autonomous mission's course. Each of the parts mentioned above is detailed in the following sections.

## V. FEATURE EXTRACTION

Aiming at an informative map constructed solely from visual sensing, a suitable representation of the recorded data is needed. It is not surprising that most pipelines use feature vectors extracted from images to describe the traversed route, given their discriminative capabilities. This characteristic extends to the visual loop closure detection task and renders it essential to select an effective visual feature encoder. The traditional choice for such a mechanism refers to hand-crafted features that are manually designed to extract specific image characteristics. Recently, however, the outstanding achievements in several computer vision tasks through deep learning have turned the scientific focus towards learned features extracted from CNN activations. A categorization of these methods is provided in Table I.

### A. Hand-Crafted Feature-Based Representation

It is shown via various experimental studies that humans can rapidly categorize a scene using only the crude global information or "gist" of a scene [177], [178]. Similarly, methods implemented upon global feature extractors describe an image's appearance holistically utilizing a single vector. Their main advantages are the compact representation and computational efficiency, leading to lower storage consumption and faster indexing while querying the database. However, hand-crafted global extractors suffer from their inability to handle occlusions, incorporate geometric information, and retain invariance over image transformations, such as those originated from the camera's motion or illumination variations. On the other hand, detecting regions-of-interest, using hand-crafted local extractors, in the image and subsequently describing them has shown robustness against transformations such as rotation, scale, and some lighting variations, and in turn, allow recognition even in cases of partial occlusions. Moreover, as the local features' geometry is incorporated, they are naturally intertwined with metric pose estimation algorithms, while their spatial information is necessary when geometrical verification is performed, as discussed in Section VII-C. In the last decade, most of the advances achieved in visual loop closure detection were based on such features. Lastly, it is the scenario wherein the robot needs to operate that drives the selection of feature extraction method. For instance, when an environment is recorded under severe viewpoint variations, a method based on local extractors would be typically preferred, while for robot applications where low computational complexity is critical, global descriptors fit better to the task. An overview of both methods is illustrated in Fig. 6.

TABLE I
METHOD'S CATEGORIZATION BASED ON THEIR *Feature Extraction* AND *Environment Representation* ATTRIBUTES

| Feature Extraction | Looking Behind | |
|---|---|---|
| | Single-image-based representation | Sequence-of-images-based representation |
| Hand-crafted global features | [97]–[108] | [109]–[116] |
| Hand-crafted local features | [117]–[139] | [140]–[144] |
| Learned image-based features | [145]–[149] | [150]–[156] |
| Learned pre-defined region-based features | [157]–[163] | - |
| Learned extracted region-based features | [164]–[169] | [170]–[172] |
| Learned extracted simultaneously image-based and region-based features | [173], [174] | - |

*1) Global Features:* Oliva and Torralba proposed the most recognized global descriptor, widely known as Gist [179]–[181], inspiring several loop closure detection pipelines [97]–[100]. A compact feature vector was generated through image gradients extracted from Gabor filters, ranging in spatial scales and frequencies. Following the Gist's success, Sünderhauf and Protzel achieved to detect loops through BRIEF-Gist [101], a global model of BRIEF (BRIEF stands for binary robust independent elementary features [182]) local descriptor to represent the entire image. Likewise, using the speeded-up robust features (SURF) method [176], a global descriptor called WI-SURF was proposed in [89]. In [183], the authors showed that when applying disparity information on the local difference binary (LDB) descriptor [102], [103], failures due to perceptual aliasing could be reduced.

Besides, another series of techniques for describing images globally is based on histogram statistics. Different forms, *e.g.,* color histograms [184], histogram-of-oriented-gradients (HOG) [104], [110], or composed receptive field histograms [185], were adopted. HOG [175], which is the most frequently used technique, calculates every pixel's gradient and creates a histogram based on the results (see Fig. 6a), while pyramid-of-HOG (PHOG) describes an image via its local shape and its spatial layout [186]. A differentiable version of HOG was introduced in [187]. Customized descriptors, originated from downsampled patch-based representations [111], constitute another widely utilized description method [113], [114]. A global descriptor derived from principal component analysis (PCA) was employed in [105].

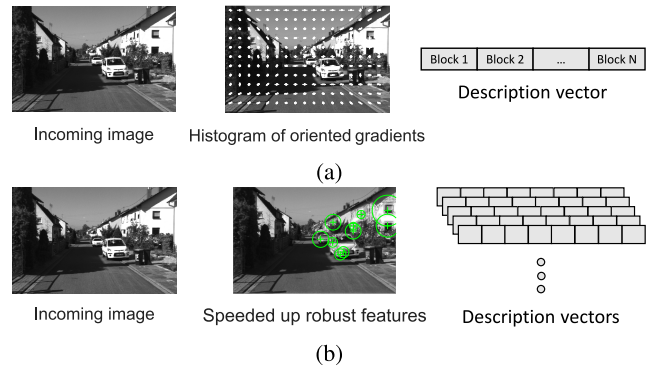

Fig. 6. Instances of hand-crafted feature descriptors, both (a) global (based on the entire image) and (b) local (based on regions-of-interest), extracted from the incoming image. (a) Whole image descriptors process each block in the image regardless of its context, *e.g.,* the histogram-of-oriented-gradients (HOG) [175]. (b) Local features, like the speeded-up robust features (SURF) [176], are indicated in salient parts of the image and subsequently described. This way, a camera measurement is represented by the total of samples.

As opposed to many of the above mentioned global description techniques, *viz.,* Gist and HOG, which are able to encode viewpoint information through concatenation of grid cells, the model of visual bag of words (BoW) describes the incoming image holistically, retaining invariant viewpoint information. In particular, this model, which was initially developed for language processing and information retrieval tasks [188], allows the images' description as an aggregation of quantized local features, *i.e.,* "visual words" [189]. More specifically, local features are classified according to a unique database, known as "visual vocabulary," generated through unsupervised density estimation techniques [190] over a set of training descriptors (either real-valued [106]–[108] or binary ones [93], [115], [141], [143]). An overview of this process is illustrated in Fig. 7. However, as several visual words may occur more frequently than others, the term-frequency inverse-document-frequency (TF-IDF) scheme [191] has been adopted to weight each database element. This way, each visual word is associated with a product proportional to the number of occurrences in a given image (term frequency) and inversely proportional to its instances in the training set (inverse document frequency). Then, every image is represented via a vector of all its TF-IDF word values [192]. Fisher Kernels [193], [194] refine the visual BoW model via fitting a Gaussian mixture over the database entries and the local features. At the same time, VLAD (VLAD stands for vector of locally aggregated descriptors [195], [196]) concatenated the distance vectors between each local feature and its nearest visual words leading to improved performance results in the cost of increasing the memory footprint.

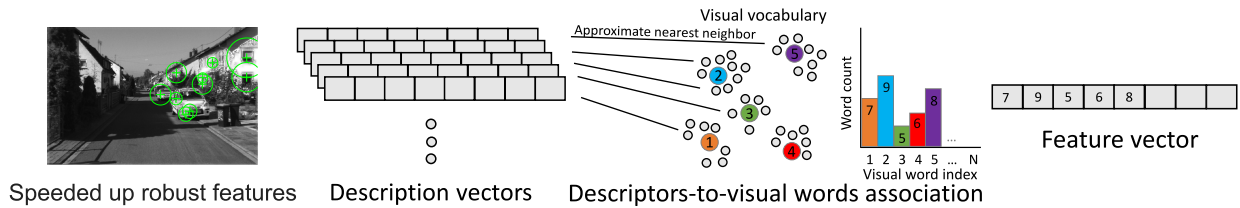

Fig. 7. The visual bag of words model based on a previously trained visual vocabulary. Speeded-up robust features [176] are extracted from regions-of-interest in the incoming image, and subsequently, their descriptors are connected with the most similar visual word in the vocabulary. The output vector ($1 \times N$ dimension, where N corresponds to the vocabulary's size) is a feature vector which represents the frequency of each visual word included in the camera data.

*2) Local Features:* Historically, the most acknowledged method for extracting local features is the scale-invariant feature transforms (SIFT) [197]. Based on the difference-of-gaussian (DoG) function, regions-of-interest are detected, while HOG computes their neighborhood's description. SURF (see Fig. 6b), inspired by SIFT, proposes a faster extraction version using an approximation of the determinant of Hessian blob detector for identifying regions-of-interest and the sum of the Haar wavelet response around each point for feature description. CenSurE [198], a lightweight equivalent of SURF, detects regions-of-interest using center-surrounded filters across multiple scales of each pixel's location. KAZE [199] detects and describes 2D features in a nonlinear scale space by means of nonlinear diffusion filtering, demonstrating improved feature quality. However, it also induces higher computationally complexity. As the research community moved towards the binary description space, various feature extractors were developed offering similar SIFT and SURF performance; yet, exhibiting reduced complexity and memory requirements. Most of them extended BRIEF, which uses simple intensity difference tests to describe regions-of-interest, by incorporating descriptiveness and invariance to scale and rotation variations, such as LDB, ORB [200], BRISK [201], FREAK [202], and M-LDB [203]. Moreover, several local extractors used geometrical cues, such as line segments [204] or integrated lines and points, into a common descriptor [205], aiming to cope with region-of-interest detection in low-textured environments.

When directly describing images using local extractors, a massive quantity of features is created [206]. This dramatically affects the system's performance, mainly when real-valued features are used [118]. Different loop closure detection pipelines partially reduce their quantity by selecting the most informative ones [117], [136], or utilizing binary descriptors to avoid such cases [119], [129]. Although the utilization of visual BoW is an efficient technique for detecting loop closures when local features are adopted, two weaknesses are presented. First, the visual vocabulary is typically generated a priori from training images and remains constant during navigation, which is practical; however, it does not adapt to the operational environment's attributes, limiting the overall loop closure detection performance. Secondly, vector quantization discards the geometrical information, reducing the system's discriminative nature, primarily in perceptual aliasing cases. Consequently, several approaches address these limitations incrementally, *i.e.,* along the navigation course, to generate the visual vocabulary [128]. This concept was introduced by Filliat [120], assuming an initial vocabulary that was gradually increased as new visual features were acquired. Similarly, Angeli *et al.* [46], [121] merged visual words through a user-defined distance threshold. Nevertheless, most incremental vocabularies (either using real valued-based [122], [124]–[127], [130] or binary descriptors [131]–[133], [135], [139]) are based on the descriptors' concatenation from multiple frames to obtain a robust representation of each region-of-interest.

### B. Learned Feature-Based Representation

CNN is a concept introduced by LeCun *et al.* in the late '80s [207], [208]. Its deployment efficiency is directly associated with the size and quality of the training process,[1] which generally constitute practical limitations [209]. However, its recent successes in the computer vision field are owed to a combination of advances in GPU computational capabilities and large labeled datasets [210]. The remarkable achievements in image classification [210], [211] and retrieval tasks [212], [213] are owed to the capability of CNNs to learn visual features with increased levels of abstraction. Hence, it was reasonable to expect that the robotics community would experiment with learned feature vectors as the loop closure detection's backbone is oblivious to the type of descriptions used.

A fundamental question is how a trained CNN generates visual representations. To answer this, we need to consider the four following paradigms that achieve feature extraction through different processes: *1)* the whole image is directly fed into a network, and the activations from one of its last hidden layers are considered as the image's descriptor [149], [215]–[217]; *2)* specific image regions are introduced to the trained CNN, while the respective activations are aggregated to form the final representation [218]–[222]; *3)* the CNN receives the whole image, and via the direct extraction of distinct patterns based on the convolutional layers' responses, the most prominent regions are detected [214], [223]–[226]; *4)* the CNN receives the whole image, and simultaneously predicts global and local descriptors [173], [174]. An illustrative paradigm is shown in Fig. 8. Generally, representing images globally using techniques from the first category shows reduced robustness when effects, such as partial occlusion or severe viewpoint variations, are presented. Image features emerging from the second category usually cope with viewpoint changes more effectively but are computationally costly since they rely on external landmark detectors. Finally, features that emerge from the third category leverage both variations, *i.e.*, viewpoint and appearance, while features from the last category resulting in significant run-time savings.

[1]For place recognition, large-scale annotated datasets from a multitude of environments, such as a comprehensive set of urban areas, are needed.

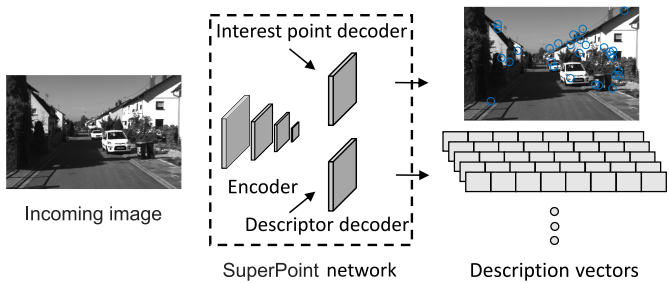

Fig. 8. A representative example of a fully-convolutional network that jointly extracts points of interest and their descriptors from an image [214].

*1) Image-Based Features:* Chen *et al.* [145] were the first to exploit learned features extracted from all layers of a trained network [215] for object recognition to detect similar locations. However, subsequent studies showed that the utilization of intermediate representations with and without the CNN's fully connected layers could offer high performances [146], [148] and rich semantic information [227], [228]. Other recent contributions provided helpful insights for better understanding the complex relationship between network layers and their features visualization [229], [230]. Since then, different architectures with slight modifications have been developed and used for visual loop closure detection [147], [151], [153]. Inspired by the success of VLAD, NetVLAD [216] was proposed as a trainable and generalized layer that forms an image descriptor via combining features, while the spatial pyramid-enhanced VLAD (SPE-VLAD) layer improved VLAD features by exploiting the images' spatial pyramid structure [149]. In all of the above, powerful network models were utilized as the base architecture, *viz.*, AlexNet [210], VGG [231], ResNet [232], Inception [233], DenseNet [234], and MobileNet [235].

*2) Pre-Defined Region-Based Features:* Compared to the holistic approaches mentioned above, another line of works relied on detecting image landmarks, *e.g.*, semantic segmentation and object distribution, originated from image patches to describe the visual data [157]–[163]. More specifically, in [212], learned local features extracted from image regions were aggregated in a VLAD fashion, while descriptors from semantic histograms and HOG were concatenated in a single vector in [158]. VLASE [159] relied on semantic edges for the image's description [236]. In particular, pixels which lay on a semantic edge were treated as entities of interest and described with a probability distribution (as given by CNN's last layer). The rest of the description pipeline was similar to VLAD. Similarly, Benbihi *et al.* presented the WASABI image descriptor for place recognition across seasons built from the image's semantic edges' wavelet transforms [163]. It represented the image content through its semantic edges' geometry, exploiting their invariance concerning illumination, weather, and seasons. Finally, a graph-based image representation was proposed in [162], which leveraged both the scene's geometry and semantics.

*3) Extracted Region-Based Features:* The idea of detecting salient regions from late convolutional layers instead of using a fixed grid and then describing these regions directly as features have achieved impressive results [164]–[167]. Regions of maximum activated convolutions (R-MAC) used max-pooling on cropped areas of the convolutional layers' feature maps to detect regions-of-interest [223]. Neubert and Protzel presented a multiscale super-pixel grid (SP-Grid) for extracting features from multiscale patches [224]. Deep local features (DELF) combined traditional local feature extraction with deep learning [225]. Regions-of-interest were selected based on an attention mechanism, while dense, localized features were used for their description. SuperPoint [214] and D2-net [226] were robust across various conditional changes. By extracting unique patterns based on the strongest convolutional layers' responses, the most prominent regions were selected in [164]. Multiple learned features were then generated from the activations within each spatial region in the previous convolutional layer. This technique was additionally extended by a flexible attention-based model in [165]. Garg *et al.* built a local semantic tensor (LoST) from a dense semantic segmentation network [170], while a two-stage system based on semantic entities and their geometric relationships was shown in [167]. Region-VLAD (R-VLAD) [166] combines a low-complexity CNN-based regional detection module with VLAD. DELF was recently extended by R-VLAD via down-weighting all the regional residuals and storing a single aggregated descriptor for each entity of interest [237].

*4) Extracted Simultaneously Image-Based and Region-Based Features:* Aiming to bridge the gap between robustness and efficiency, an emerging trend in the line of learned features combines the advances in the aforementioned fields to jointly estimate global and local features [173], [174]. Hierarchical feature network (HF-Net) [173], a compressed model trained in a flexible way using multitask distillation, constitutes a fast; yet, robust and accurate technique for localization tasks. In [174], the authors unify global and local features into a single model referred as DELG (stands for DEep Local and Global features). By combining generalized mean pooling for global features and attentive selection for local features, the network enables accurate image retrieval.

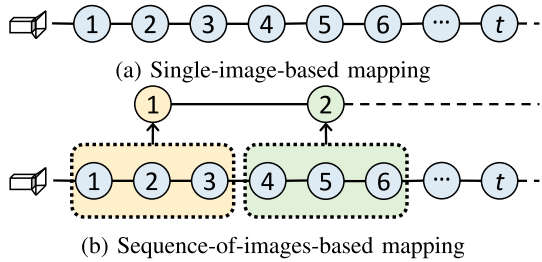

Fig. 9. Depending on their trajectory mapping, appearance-based systems are divided into two main categories, namely (a) single-image-based and (b) sequence-of-images-based. Methods of the former category represent each image in the database as a distinct location, while the latter category's schemes generate sequences, i.e., groups of individual images, along the navigation course. The observations included in each of these sequences, also referred to as sub-maps, typically consist of common visual data.

## VI. Looking Behind

As mentioned earlier, visual localization and loop closure detection are quite similar tasks. They share the primary goal of finding the database's most alike view, but for loop detection, all images acquired during the robot's first visit to a given area are treated as the reference set for a query view. As the system processes the sensory input data, it incrementally generates the internal map, i.e., database, which plays a vital role in the subsequent steps for location indexing and confidence estimation about its current position. Depending on how the robot maps the environment, visual loop closure detection pipelines are distinguished into single-image-based and sequence-of-images-based. Frameworks of the first category seek the most identical view in the robot's route, while techniques belonging in the second category look for the proper location between sub-maps, i.e., groups of individual images. This section's remainder briefly describes representative approaches by distinguishing them based on how they map the trajectory and how the system searches the database for potential matches.

### A. Environment Representation

Single-image-based mapping is the most common scheme for visual loop closure detection. During navigation, the extracted visual features from each input image are associated with a specific location (see Fig. 9a). When the off-line visual BoW model is used, the map is formulated as a set of vectors denoting visual words at each location [106]. Otherwise, a database of descriptors indexed according to their extracted location is built [118], [136].

In contrast to the conventional single-image-based methods, various frameworks use image-sequence partitioning (ISP) techniques to define group-of-images along the traversed route, which are defined as smaller sub-maps [108], [116], [238]–[240], as illustrated in Fig. 9b. These techniques either use the single-image-based representation for their members [111], or they describe each submap through sequential descriptors [144], [168], [171], [172]. However, many challenges emerge when splitting the map into sub-maps, such as optimal size, sub-map overlapping throughout database searching, and uniform semantic map definition [80]. SeqSLAM [111], the most acknowledged algorithm in sequence-of-images-based mapping, has inspired a wide range of authors since its first introduction [110], [112], [142], [150], [152], [154]–[156], [241]–[243]. The multitude of these pipelines, with SeqSLAM among them, uses a pre-defined quantity of images to segment the trajectory. Nevertheless, the unknown frame density, out-of-order traverses, and diverse frame separation are some of the characteristics which negatively affect the fixed-length sub-mapping methods' performance. To avoid such cases, dynamical sequence definition techniques are employed using landmarks' co-visibility properties [244]–[247], features' consistency among consecutive images [109], [113], [114], [143], temporal models [70], [156], or transition-based sub-mapping, e.g., through particle filtering [241].

### B. Location Indexing

A visual loop closure detection system must search for similar views among the ones visited to decide whether a query instance corresponds to a revisited location. Firstly, recent database images should not share any familiar landmarks with the query. This is because images immediately preceding the query are usually similar in appearance to the recent view; however, they do not imply that the area is revisited. Aiming to prevent the system from detecting false-positives, these locations are rejected based on a sliding window defined either by a timing constant [127], [130], [135], [169], [248] or environmental semantic changes [46], [90], [121], [137], [141], [143]. Methods based on the off-line visual BoW model employ the inverted indexing technique for searching, wherein the query's visual words indicate the locations that have to be considered as potential loop events. In contrast, methods that do not follow this model implement an exhaustive search on the database descriptors' space [71], [119], [130], [134], [137], [139].

## VII. Decision Making

The final step is the decision of whether the robot observes a previously mapped area or not. Different comparison techniques, which are broadly classified according to their map representation, have been proposed to quantify this confidence [4]; the first one is image-to-image, and the second is sequence-to-sequence. The former computes an individual similarity score for each database entry [106], [123], [134], which is then compared against a pre-defined hypothesis threshold to determine whether the new image is topologically connected to the older one. Otherwise, the query cannot match any pre-visited one, resulting in a new location addition to the database. On the contrary, sequence-to-sequence is typically based on the comparison of sub-maps [140]–[144]. Subsequently, loop closing image pairs are considered the groups' members with the highest similarity scores.

Moreover, to avoid erroneous detections, both temporal and geometrical constraints are employed, primarily to address perceptual aliasing conditions. Representative examples include recognizing a closed-loop only if supported by neighboring ones or if a valid geometrical transformation can be computed between the matched frames. As a final note, the resulting confidence metrics fill a square matrix whose $(i, j)$ indexes denote the similarity between images $I_i$ and $I_j$.

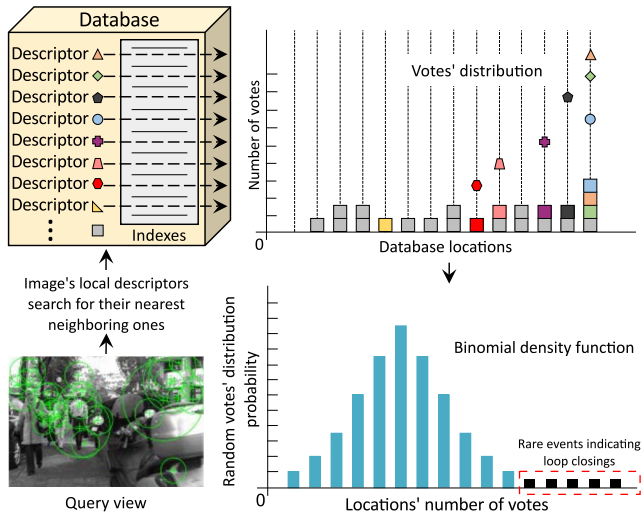

Fig. 10. As the most recently obtained image's local descriptors are extracted at query time, votes are distributed to database locations $l$ from where their nearest neighbor descriptor originates. The colored and gray cubes represent the votes casted to several locations. After the locations' polling, a voting score is received that is used to evaluate the similarity. The naïve approach is based on the number of votes (top-right); however, since thresholding the number of votes is not intuitive, more sophisticated methods, such as binomial density function [129], utilize the location's total amount of aggregated votes to compute a probabilistic score which highlights loop closure detections.

### A. Matching Locations

Sum of absolute differences (SAD), a location's similarity votes density, and Euclidean or cosine distance are the commonly used metrics employed to estimate the matching confidence between two instances. Directly matching the features extracted from two images represents a reasonable similarity measurement when global representations are used (either hand-crafted or learned). However, when local features are selected, voting schemes are selected. These techniques depend on the number of feature correspondences leading to an aggregation of votes, the density of which essentially denotes the similarity [105] between two locations. This is typically implemented by a $k$-nearest neighbor ($k$-NN) search [123], [132], [134], [244], [245]. The simple approach is to count the number of votes and apply heuristic normalization [134]; however, in these cases, thresholding is not intuitive and varies depending on the environment. Rather than naïvely scoring the images based on their number of votes, Gehrig *et al.* [129] proposed a novel probabilistic model originated from the binomial distribution (see Fig. 10) [130], [136], [137]. By casting the problem into a probabilistic scheme, the heuristic parameters' effect is suppressed, providing an effective score to classify matching and non-matching locations, even under perceptual aliasing conditions. Over the years, probabilistic scores were used to enhance the system's confidence [241]. The Dempster-Shafer probability theory, which models ignorance without prior knowledge of the environment, was introduced by Savelli and Kuipers [249]. Similar to the Bayes approach discussed in [11], [15], later works followed the Bayesian filtering scheme to evaluate loop closure hypotheses [46], [90], [100], [106], [107], [121], [127], [135], [138].

Each of the techniques mentioned above can be efficiently adopted in sequence-of-images-based methods (*e.g.,* SeqSLAM [111], HMM-SeqSLAM [250], ABLE-M [251], S-VWV [115], MCN [252]). By comparing route segments rather than individual camera observations, global representations are able to provide outstanding results through the utilization of relatively simple techniques. As shown in SeqSLAM, to evaluate the locations' similarity, the SAD metric is used between contrast-enhanced, low-resolution images avoiding this way the need for key-points extraction. For a given query image, comparisons between the local query sub-map and the database are performed. The likelihood score is the maximum sum of normalized similarity scores over the length of pre-defined constant velocity assumptions, *i.e.,* alignments among the query sequence and the database sequence images. This process is inspired by speech recognition and is referred to as continuous dynamic time warping (DTW) [253]. Alignment is solved by finding the minimum cost path [254], while dynamic programming [255], graph-based optimization [256]–[259], or the incorporation of odometry information [142] strengthens its performance [250]. To improve the systems' performance, frameworks based on dynamic adjustment of the sequence length are also proposed that leverage feature matching [113], [114], GPS priors [251], or modeling the area hypotheses over different length assumptions [260], [261].

### B. Exploiting the Temporal Consistency

In robot navigation, unlike large-scale image retrieval or classification tasks, where images are disorganized, sensory measurements are captured sequentially and without time gaps [262], [263]. Most pipelines pay a high price for indicating a loop closure, but there is minor harm if one is missed since many chances in the following images are afforded due to the existing temporal continuity. Every sequence-of-images-based mapping technique leverages the sequential characteristic of robotic data streams aiming to disambiguate the boisterous single-image-based matching accuracy. The temporal consistency constraint, which is mainly adopted when single-image-based mapping is used, filters out inconsistent loop closures through heuristic methods (*e.g.,* continuous loop hypothesis before a query is accepted [130], [132], [137], [248]) or more sophisticated ones (*e.g.,* the Bayesian filter [100], [106], [121], [127], [135], [138]).

### C. Is the Current Location Known? The Geometrical Verification

After data association, a geometrical verification check is often implemented based on the spatial information provided in local features, either hand-crafted or learned ones, by computing a fundamental/essential matrix or other epipolar constraints [72], [78], [82], [84], [106], [119], [130], [135]–[137], [141], [237], [264]–[266]. Typically, it is performed using some variation of the RANSAC algorithm and additionally provides the relative pose transformation if a successful correspondence is found [267]. Moreover, a minimum number of RANSAC inliers has to be satisfied for a loop to be confirmed [268].

When a stereo camera rig is used [94], a valid spatial transformation between the two pairs of matching images is computed through the widely used iterative closest point (ICP) algorithm for matching 3D geometry [269]. Given an initial starting transformation, ICP iteratively determines the transformation among two point clouds that minimizes their points' error. Still, a high computational cost accompanies the matching process when the visual and the 3D information are combined [106]. As a final note, geometrical verification is based on the spatial information of hand-crafted local features. Typically, a system that uses single vector representations (either global or visual BoW histograms) needs to further extract local features, adding more complexity to the algorithm.

## VIII. Benchmarking

In order to benchmark a given loop closure detection approach, three major components are mainly used: the datasets, the ground truth information, and the evaluation metrics [38]. Accordingly to the case under study, a variety of datasets exist in the literature. The ground truth is typically formed in the shape of a boolean matrix whose columns and rows denote observations recorded at different time indices $(i, j)$. Hence, the 1 indicates a loop closure event between instances $i$ and $j$ and 0 otherwise. This matrix, together with the similarity one, is used to estimate how the system performs. Typically, the off-diagonal high-similarity elements of the generated similarity matrix indicate the locations where loops are closed. Finally, the chosen evaluation metric is the last component needed for measuring the performance.

### A. Evaluation Metrics

The relatively recent growth of the field has led to the development of a wide variety of datasets and evaluation techniques, usually focusing on precision-recall metrics [282]. These are computed from a loop closure detection algorithm's outcome: the correct matches are considered true-positives, whereas the wrong ones as false-positives. In particular, whether or not two images originate from the same area determines a positive or negative loop closure event, respectively. Moreover, the outcome of a loop closure detection algorithm can be characterized as true or false based on the information provided by the ground truth. Therefore, we end up with four different detection labels: true-positive, false-positive, true-negative, and false-negative. On the one hand, a true-positive instance is regarded as any database entry identified by the algorithm within a small radius from the query's location, while false-positives refer to incorrect detections that the system may produce and lie outside this range. On the other hand, false-negatives are the loops that had to be detected; yet, the system could not identify them, while true-negatives are those that the system correctly rejects.

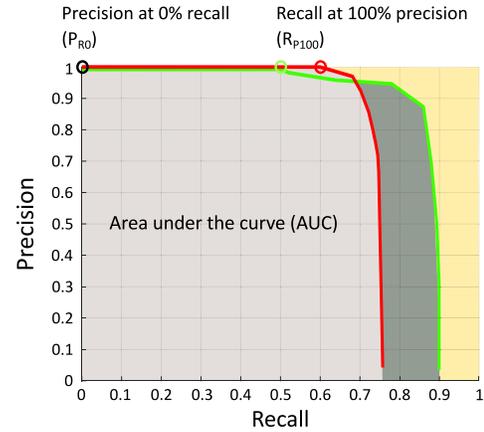

Fig. 11. An illustrative example of two hypothetical precision recall curves monitoring a method's performance. A curve is extracted by altering one of the system's parameter. The highest possible recall score for a perfect precision ($R_{P100}$), which is the most common indicator for measuring the system's performance, is shown by the red and green cycles. The precision at minimum recall ($P_{R0}$) is depicted by the black cycle, while the gray color areas denote the area under the curve. At a glance, the two curves suggest that the red curve is better than the green one. Indeed, the corresponding metrics, i.e., the $R_{P100}$ at 0.6 and the expected precision at 0.8, confirm that the red curve denotes improved performance even if the area under the curve is larger in the green curve.

Thus precision is defined as the number of accurate matches (true-positives) overall system's detections (true-positives plus false-positives):

$$\text{Presicion} = \frac{\text{True-positives}}{\text{True-positives} + \text{False-positives}}, \quad (1)$$

whereas recall denotes the ratio between true-positives and the whole ground truth (sum of true-positives and false-negatives):

$$\text{Recall} = \frac{\text{True-positive}}{\text{True-positive} + \text{False-negatives}}. \quad (2)$$

A precision-recall curve shows the relationship between these metrics and can be obtained by varying a system's parameter responsible for accepting of a positive match, such as the loop closure hypothesis threshold [283]. The area under the precision-recall curve (AUC) [284] is another straightforward metric for indicating the performance [156], [164]. Its value ranges between 0 and 1; yet, any information with respect to the curve's characteristics is not retained in AUC, including whether or not the precision reaches 100% at any recall value [285]. The average precision is also helpful when the performance needs to be described by a single value [286]. Generally, a high precision across all recall values is the main goal for a loop closure detection system, and average precision is capable of capturing this property. However, the most common performance indicator for evaluating a loop closure detection pipeline is the recall at 100% precision ($R_{P100}$). It represents the highest possible recall score for a perfect precision (i.e., without false-positives), and it is a critical indicator since a single false-positive detection can, in many cases, cause a total failure for SLAM. Nevertheless, $R_{P100}$ cannot be determined when the generated curves are unable to reach a score for 100% precision. To overcome this problem, the extended precision (EP) metric is introduced as: $EP = (P_{R0} + R_{P100})/2$ [287]. EP summarizes a precision-recall curve through the combination of two of its most significant features, namely, precision at minimum recall ($P_{R0}$) and $R_{P100}$, into a comprehensible value. In a similar manner, the recall score for 95% precision [288] is another metric for assessing visual loop closure detection systems, as a small number of fault detections can be further validated through SLAM's back-end optimization techniques, *e.g.*, via robust pose-graph optimization [289]–[292]. In Fig. 11, a representative example of two hypothetical precision recall curves is given. As shown, each depicted evaluation metric indicates that the red curve produces better performance ($R_{P100} = 0.6$ and $EP = 0.8$) than the green one; although, the area under the curve is larger in the latter hypothesis.

TABLE II
DESCRIPTION OF LOOP CLOSURE DETECTION DATASETS WITH *fixed* ENVIRONMENTAL CONDITIONS

| Dataset | Sensor characteristics | Characteristics | Image resolution & frequency | # Frames | Traversed distance | Ground truth |
|---|---|---|---|---|---|---|
| KITTI (course 00) [270] | Stereo, Gray, frontal | Outdoor, urban, dynamic | $1241 \times 376$, 10 Hz | 4551 | $\sim$12.5 km | ✓ |
| KITTI (course 02) [270] | Stereo, Gray, frontal | Outdoor, urban, dynamic | $1241 \times 376$, 10 Hz | 4661 | $\sim$13.0 km | ✓ |
| KITTI (course 05) [270] | Stereo, Gray, frontal | Outdoor, urban, dynamic | $1241 \times 376$, 10 Hz | 2761 | $\sim$7.5 km | ✓ |
| KITTI (course 06) [270] | Stereo, Gray, frontal | Outdoor, urban, dynamic | $1241 \times 376$, 10 Hz | 1101 | $\sim$3.0 km | ✓ |
| Lip6O [121] | Mono, color, frontal | Outdoor, urban, dynamic | $240 \times 192$, 1 Hz | 1063 | $\sim$1.5 km | ✓ |
| Lip6I [121] | Mono, color, frontal | Indoor, static | $240 \times 192$, 1 Hz | 388 | $\sim$0.5 km | ✓ |
| City Centre [106] | Stereo, color, lateral | Outdoor, urban, dynamic | $1024 \times 768$, 7 Hz | 1237 | $\sim$1.9 km | ✓ |
| New College [106] | Stereo, color, lateral | Outdoor, static | $1024 \times 768$, 7 Hz | 1073 | $\sim$2.0 km | ✓ |
| Eynsham [271] | Omnidirectional, gray | Outdoor, urban, rural | $512 \times 384$, 20 Hz | 9575 | $\sim$70.0 km | ✓ |
| New College vision suite [2] | Stereo, gray, frontal | Outdoor, dynamic | $512 \times 384$, 20 Hz | 52480 | $\sim$2.2 km | ✓ |
| Ford Campus (course 02) [272] | Omnidirectional, color | Outdoor, urban | $1600 \times 600$, 8 Hz | 1182 | $\sim$ 10km | ✓ |
| Malaga 2009 Parking 6L [273] | Stereo, color, frontal | Outdoor, static | $1024 \times 768$, 7 Hz | 3474 | $\sim$ 1.2km | ✓ |
| EuRoC Machine Hall 05 [274] | Stereo, gray, frontal | Indoor, static | $752 \times 480$, 20 Hz | 2273 | $\sim$ 0.1km | ✓ |

TABLE III
DESCRIPTION OF LOOP CLOSURE DETECTION DATASETS WITH *Changing* ENVIRONMENTAL CONDITIONS

| Dataset | Sensor characteristics | Characteristics | Image resolution & frequency | Ground truth |
|---|---|---|---|---|
| Symphony Lake [275] | Omnidirectional, color, frontal | Outdoor, changing, static | $704 \times 480$, 10 Hz | ✓ |
| SFU Mountain [276] | Stereo & mono, color & gray, frontal | Outdoor, changing, static | $752 \times 480$, 30 Hz | ✓ |
| Gardens Point [277] | Mono, color, frontal | Outdoor, changing | $1920 \times 1080$, 30 Hz | ✓ |
| St. Lucia [278] | Mono, color, frontal | Outdoor, slightly changing, dynamic | $640 \times 480$, 15 Hz | ✓ |
| Oxford RobotCar [279] | Trinocular stereo, color, frontal | Outdoor, changing, highly dynamic | $1280 \times 960$, 16 Hz | ✓ |
| Nordland [242] | Mono, color, frontal | Outdoor, changing, static | $1920 \times 1080$, 25 Hz | ✓ |
| Mapillary [218] | User submitted | Outdoor, viewpoint, dynamic | N/A | ✓ |
| Synthia [280] | Synthetic | Synthetic, viewpoint, dynamic | $960 \times 720$, N/A Hz | ✓ |
| Lagout [20] | Synthetic | Outdoor, viewpoint, dynamic | $752 \times 480$, N/A Hz | ✓ |
| Corvin [20] | Synthetic | Viewpoint, static | $752 \times 480$, N/A Hz | ✓ |
| Newer College [281] | Stereo, color, frontal | Outdoor, changing, textureless surface | $848 \times 480$, 30 Hz | ✓ |

### B. Datasets

Most experiments are conducted on publicly available datasets, including urban environments, indoor and outdoor areas, recorded through various platforms. A renowned benchmark environment in the robotics community is the KITTI vision suite giving a wide range of trajectories with accurate odometry information and high-resolution image properties (for both image size and frame rate) [270]. Courses 00, 02, 05, and 06 are mainly used since they present actual loop closures compared to the rest ones. The incoming visual stream is captured via a camera system that is placed on a car. However, when monocular loop closure detection systems are proposed, only one camera stream is considered. The authors in [183] manually obtained the corresponding ground truth based on the dataset's odometry data. Early studies were based on the Lip6 Outdoor and Indoor image-sequences recorded by a handheld camera facing many loop closures in an outdoor urban environment and a hotel corridor, respectively [121]. Both are considered challenging due to the sensor's low frame rate and resolution. They contain their own ground truth information concerning the related loop closure events. City Centre, New College, and Eynsham are three datasets extensively utilized in visual SLAM and, in particular, to evaluate loop closure detection pipelines [106], [271]. The first two above were collected by a robotic platform with two cameras positioned on the left and right sides while moving through outdoor urban environments with consistent lighting conditions. Eynsham is a 70 km urban dataset consisting of two 35 km traverses. It was recorded by a Ladybug 2 camera providing panoramic images captured at 7 m intervals. Ground truth information in terms of actual loop closures is also given. Another widely used version of the New College dataset was later presented in [2]. Ford Campus is another collection of several panoramic images [272]. A considerable quantity of loop closure examples exists in both cases. Malaga 2009 Parking 6L [273] was recorded at an outdoor university campus parking lot via the stereo vision system of an electric buggy-typed vehicle. Finally, the EuRoC Machine Hall 05 is part of the EuRoC Micro Aerial Vehicle (MAV) dataset [274] and presents fast velocity changes and various loop examples with minor illumination variations. Cameras placed on a MAV obtain the respective visual data with a high acquisition frame rate. In Fig. 12, representative instances from a dataset designed explicitly for assessing loop closure detection techniques are presented, while in Table II an overview is given.

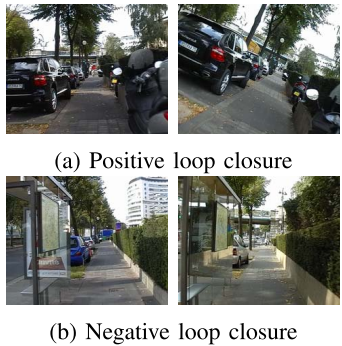

Fig. 12. Representative positive and negative instances of Lip6 Outdoor [121].

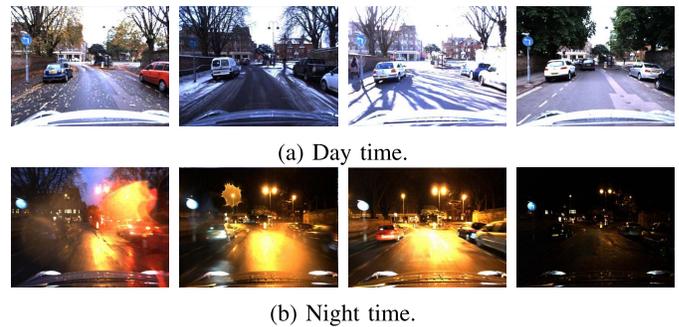

Fig. 13. Example images from the Oxford RobotCar dataset [279] for both (a) day-time and (b) night-time conditions. From left to right: Autumn, winter, spring, and summer. Within long-term and large-scale SLAM autonomy, detections need to be successful despite significant variations in the images' context, such as different illumination conditions, *e.g.*, day and night, or year seasons.

Of particular challenge are datasets captured over multiple seasons as their appearance is modified due to different weather conditions, sun position, and vegetation state (*e.g.,* Symphony Lake dataset [275]). SFU Mountain provides multiple trajectories of a mobile robot in a semi-structured woodland [276], while the Gardens Point Dataset contains three traverses of the Queensland University of Technology [277]. The first two are captured during daytime by walking on the two opposite sides of the walking path (lateral viewpoint), while the last one throughout the night. The respective sequences are synchronized; thus, ground truth is structured as frame correspondences. The St. Lucia dataset [278] comprises images recorded from a selection of streets at five different periods of a day over two weeks. It only contains images from suburb environments, and the appearance variations of each place are minor. Over 100 traverses with various weather (*e.g.,* direct sun, overcast) and illumination (*e.g.,* day, night) conditions are provided in Oxford RobotCar [279]. Several challenges of pose and occlusions, such as pedestrians, vehicles, and bicycles, are included in the recorded sequences. The Nordland dataset [242] consists of 10 hours of video footage covering four times a 728 km ride in northern Norway, one for each season. It has 37,500 images and is a highly acknowledged dataset for studying seasonal changes in natural environments. However, it does not present any variation concerning the viewpoint between the platform's paths. In the Mapillary dataset, three sequences were recorded in Berlin city streets, presenting severe viewpoint changes with moderate conditional variations [218]. Synthia [280] is a synthetically created dataset containing trajectories in city-like environments throughout spring and winter. 959 and 947 are the total of query and reference instances, respectively. Consisting of several flybys around buildings, Lagout and Corvin are also two synthetic environments [20]. Lagout sequences 0° and 15° are used as reference and query datasets, respectively, to test visual place recognition techniques under moderate viewpoint changes. In a similar manner, Corvin's loops, which are recorded at ground level, are utilized to assess visual place recognition methods under tolerant viewpoint variances Ground truth data regarding the included loop closure events for Lagout and Corvin are made available by their authors. The Newer College dataset [281] contains challenging image-sequences with aggressive shaking, fast motion, textureless surface, and rapid lighting change through a stereo camera system in New College, Oxford. A synopsis of these datasets is given in Table III.

## IX. NEW CHALLENGES: LONG-TERM OPERATION

The main objective of any loop closure detection pipeline is to facilitate robust navigation for an extended period and under a broad range of viewing situations. Within long-term and large-scale SLAM autonomy, previously visited locations in dynamic environments need to be recognized under different day periods and scenes with changeable illumination and seasonal conditions [86], [183], [251], [293]. As a result, it becomes increasingly difficult to match two images, mainly since such variations affect the image appearance significantly (Fig. 13). Furthermore, extreme viewpoint variations lead to severe perspective distortions and low overlap between the query and the database frames.

Another critical aspect in long-term applications is the storage requirements needed to map the whole environment effectively. The majority of approaches scale linearly to the map's size (at best). Consequently, there has been much interest in developing compact appearance representations so as to demonstrate sub-linear scaling in computational complexity and memory demands. These techniques typically trade off memory usage with detection performance, or vice versa, for achieving computational efficiency.

### A. Dynamic Environments

During navigation in a changing environment, the topological information about the robot's relative movement becomes more important as noise from the sensory inputs is accumulated to an overwhelming degree [294], [295]. Early works exploited the topological information through sequence matching [111], [250], or network flows [243]. However, their output is still dependent on their visual representations' quality since the utilized hand-crafted features were not distinctive enough so as to form a genuinely reusable map [74], [278], [296]. On the contrary, representations provided by deep learning techniques show promising results on applications with challenging conditional and viewpoint changes [162], [217], [224], [297], [298]. More specifically, deep learning approaches can be utilized to either construct description features with increased robustness to perceptual changes [104], [111], [278] or to predict and negate the effect of appearance variations [211], [243], [299]–[301]. It is also worth noting that for both the above cases, networks that are previously trained for semantic place classification [302] outperform the ones designed for object recognition when applied for place recognition under severe appearance changes [228]. Moreover, as the vast majority of the approaches and datasets assume a static environment which limits the applicability of visual SLAM in many relevant cases, such as intelligent autonomous systems operating in populated real-world environments [303], [304]. Detecting and dealing with dynamic objects is a requisite to estimate stable maps, useful for long-term applications. If the dynamic content is not detected, it becomes part of the 3D map, complicating tracking or localization processes.

*1) Robust Visual Representations:* Such techniques are mainly based on a single global descriptor. SeqSLAM constitutes a representative example for this category, and it is extensively utilized to recognize similar locations under drastically different weather and lighting conditions by using sequence-of-images-based matching. A series of subsequent works have been developed following the same architecture [243], [305], that does not adopt learned features as their representation mechanism. Among the different variants, the gist-based pipeline [100] compares the learning-based ones [306], [307]. Another approach by Vidas and Maddern [75] utilized two different visual vocabularies by combining SURF-based visual words from the visible and infrared spectrum. Their results showed that hand-crafted features could not achieve high performances in complicated dynamic environments; however, the infrared data were more robust to extreme variations. On the other hand, techniques which are built upon learned features typically demand an extensive labeled training set [163], [164], [214], [226], [286], [308]; however, there exist some exceptions that do not require environment-specific learning samples [156], [309].

*2) Learning and Predicting the Appearance Changes:* These methods require labeled training data, such as matched frames from the exact locations under different conditions [171], [299], [301], [310], [311]. In [312], an average description of images was learned, *viz.,* a vector of weighted SIFT features. Their system was trained in summer and winter environments looking for valuable features capable of recognizing places under seasonal changes. The features that co-occurred in each image taken at different times of the day were combined into a unique representation with identifiable points from any point of view, irrespective of illumination conditions [109]. Similarly, matching observations with significant appearance changes was achieved using a support-vector machine (SVM) classifier to learn patch-based distinctive visual elements [104], [313]. This approach yields excellent performance but has the highly restrictive requirement that training must occur in the testing environment under all possible environmental conditions. The authors in [305] learned how the appearance of a location changes gradually, while Neubert *et al.* [299] constructed a map based on visual words originated from two different conditions. A super-pixel dictionary of hand-crafted features specific for each season was built in [311] by exploiting the seasonal appearance changes' repeatability. Using change-removal, which is similar to dimensionality reduction, showed that by excluding the less discriminative elements of a descriptor, an enhanced performance could be achieved [19], [314]. Another way to tackle such challenges was based on illumination-invariant image conversions [211], [300], [315], and shadow removal [316]–[318]. The former transferred images into an illumination invariant representation; however, it was shown that the hypothesis of a black-body illumination was violated, yielding poor results [211]. Shadow removal techniques were used to obtain invariant illumination images independent of the sun's positions.

Lategahn *et al.* [319] were the first to study how the CNNs can be used for learning illumination invariant descriptors automatically. A network selected the subset of the visual features, which were consistent between two different appearances of the same location [320]. Exploiting the visual features extracted from ConvNet [218], a graph-based visual loop detection system was proposed in [263], while a BoW for landmark selection was learned in [321]. Modifying images to emulate similar query and reference conditions is another way to avoid addressing the descriptors for condition invariance. The authors in [322] learned an invertible generator, which transformed the images to opposing conditions, *e.g.,* summer to winter. Their network was trained to output synthetic images optimized for feature matching. Milford *et al.* [323] proposed a model to estimate the corresponding depth images that are potentially condition-invariant.

*B. Viewpoint Variations*

Viewpoint changes are as critical as the appearance variations since visual data of the same location may seem much different when captured from other views [324]. The variation in viewpoint could be a minor lateral change or a much-complicated one, such as bi-directional or angular changes coupled with alterations in the zoom, base point, and focus throughout repeated traverses. Over the years, most pipelines were focused on unidirectional loop closure detections. However, in some cases, they were not sufficient for identifying previously visited areas due to bidirectional loop closures, *i.e.,* when a robot traverses a location from the opposite direction. This type of problem is crucial because solely unidirectional detections do not provide robustness in long-term navigation. Traditional pipelines, such as ABLE-P [325], identified bidirectional loops by incorporating panoramic imagery. A correspondence function to model the bidirectional transformation, estimated by a support-vector regression technique, was designed by the authors in [326] to reject mismatches.

To achieve greater viewpoint robustness, semantically meaningful mapping techniques were adopted to detect and correct large loops [146], [164], [327]. Using visual semantics, extracted via RefineNet [328], multi-frame LoST-X [170] accomplished place recognition over opposing viewpoints. Similarly, appearance invariant descriptors (*e.g.,* objects detected with CNN [145], [151], [161], [162], [166], [218], [221], [329], [330] or hand-crafted rules [224], [331]) showed that semantic information can provide a higher degree of invariability. Likewise, co-visibility graphs, generated from learned features, could boost the invariance to viewpoint changes [100], [222].

Finally, another research trend which has recently appeared tries to address the significant changes in viewpoint when images are captured from ground to aerial platforms using learning techniques. In general, the world is observed from much the same viewpoints over repeated visits in cases of ground robots; yet, other systems, such as a small UAV, experience considerably different viewpoints which demand recognition of similar images obtained from very wide baselines [20], [332]. Traditional loop closure detection systems do not usually address such scenarios; novel algorithms have been proposed in complementary areas for ground-to-air association [333]–[338].

### C. Map Management and Storage Requirements

Scalability in terms of storage requirements is one of the main issues every autonomous system needs to address within the long-term mapping. In dense maps, in which every image is considered as a node in the topological graph, the loop closure database increases linearly with the number of images [90], [106], [141], [271]. Consequently, for long-term operations that imply an extensive collection of images, this task becomes demanding not only to the computational requirements but also the system's performance. This problem is tackled through map management techniques: 1) using sparse topological maps, representing the environment with fewer nodes which correspond to visually distinct and strategically interesting locations (key-frames), 2) representing each node in a sparse map by a group of sequential and visually similar images, and 3) limiting the map's size by memory scale discretization.

*1) Key-Frame Selection:* is based on the detection of scenes' visual changes by utilizing methods developed for video compression [339]. However, the main difference between key-frame mapping and video abstraction is that the former requires the query image's localization with a previously visited location. This is vital for the system's performance since a single area might be recorded by two different locations [56]. Both locations may reach half of the probability mass, and therefore, neither attracts the threshold for successful data matching. Traditionally, the metric for deciding when to create graph nodes was typically an arbitrary one. Representative examples include the distance and angle between observations in space [79], [92], [264], [340], specific time intervals [265], [341], and a minimum number of tracked landmarks [123], [134], [342]–[344]. An illustration of a map represented by key-frames is shown in Fig. 14.

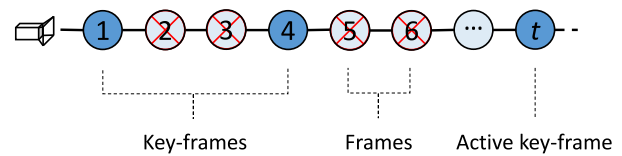

Fig. 14. Illustration of a map represented by key-frames.

*2) Representing Each Node in a Sparse Map by a Group of Sequential and Visually Similar Images:* is a well-established process that offers computational efficiency while also retaining high spatial accuracy. Techniques that fall into this category map the environment hierarchically [345]–[349] and tackle scalability through the formulation of image groups, thus reducing the database's search space [58], [59], [73], [135], [350]–[353]. Hierarchies have also been found in the mammalian brain, both in the structure of grid cells in the Hippocampus [354] and the visual cortex's pathway [355]. To limit the number of database instances, clustering [130], [240], [247], [356]–[358] or pruning [359] methods can be used and restrain map's parts which exceed a threshold based on the spatial density. Hierarchical approaches follow a two-stage process: firstly, less-intensive nodes are selected, and, next, the most similar view in the chosen node is searched [238], [360]. For instance, in [361] and [184], a hierarchical approach based on color histograms allows the identification of a matching image subset, and subsequently, SIFT features are utilized for acquiring a more precise loop closing frame within this subset. Similarly, nodes are formulated by grouping images with common visual properties, represented by an average global descriptor and a set of binary features through on-line BoW [135]. Korrapati and Mezouar [108] used hierarchical inverted files for indexing images. Exploiting the significant run-time improvements of hierarchical mapping, the authors in [173], [174], [362] achieved real-time performance using learned descriptors.

*3) Short-Memory Scale Discretization:* limit the map's size, so that loop closure detection pipelines keep a processing complexity under a fixed time constrain and satisfy the on-line requirements in long-term operations [127]. Mobile robots have limited computational resources; therefore, the map must be somewhat forgotten [127], [138], [363]–[365]. Nevertheless, this needs ignoring of locations, a technique that leads to mismatches in future missions. On the contrary, maintaining in random access memory the entire robot's visual history is also sub-optimal and, in some cases, not possible. Dayoub and Duckett [363] mapped the environment by using reference views, *i.e.,* many known points. Two specific memory time scales are included in every view: a short-term and a long-term. Frequently observed features belonging in the short-term memory advance to the long-term memory, while the ones not frequently observed are forgotten. They showed that the query view presented a higher similarity to these reference views for nine weeks [366]. Following a similar process, real-time appearance-based mapping (RTAB-MAP) [127] used short-term and long-term memory, while the authors in [246] assumed a system that includes working memory and an indexing scheme built upon the coreset streaming tree [367]. The method in [228] encoded regularly repeating visual patterns in the environment, and the management of an incremental visual vocabulary was presented in [138] based on the repetition of tracked features.

### D. Computational Complexity

In contrast to computer vision benchmarks, wherein the recognition accuracy constitutes the most crucial metric regarding performance measurement, robotics depends on flexible algorithms that can perform robustly under certain real-time restrictions. As most visual loop closure detection solutions share the concepts of feature extraction, memorization, and matching, storage and computational costs, which increase drastically with the environment size, constitute such systems' weaknesses [66], [111], [142]. Given the map management strategies mentioned in Section IX-C for large-scale operations, the main constraints to overcome are the visual information storage and the complexity of similarity computations. If one were to take the naïve approach of using an exhaustive nearest neighbor search and directly comparing all the visual features of the current robot view with all of those observed so far, the complexity of the approach would become impractical. This is due to the comparisons performed for images that do not exhibit the same context. This gets progressively less feasible as the run-time is analogous to the size of previously seen locations. Therefore, compact representations [239], [368] and hashing methods [240], [369], [370] have been explored, apart from data structure-based retrieval techniques, e.g., trees [371]–[375] and graphs [135], [376]–[379].

As the computational time of feature matching varies according to the visual feature's length, encoding the data into compact representations reduces the storage cost and simultaneously accelerates the similarity computations [380], [381]. Using the most discriminant information in high-dimensional data, Liu and Zhang [99] performed loop closure detection based on a PCA technique. They achieved to reduce the descriptor space from 960 dimensions to the 60 most discriminative ones while preserving high accuracy. Another line of frameworks adopted binary descriptors to improve computational efficiency [141] or encoded the high-dimensional vectors into compact codes, such as hashing [382]. Typical matching techniques include hashing, e.g., locality sensitive hashing (LSH) [383] or semantic hashing [384]. Although LSH does not need any pre-processing or off-line procedures [240], [370], [385]–[388], its discrete feature representations suffer from data collisions when their size is large [389]. Nevertheless, with a view to avoid data collision and achieve unique mapping, visual information is embedded in continuous instead of discrete lower-dimensional spaces [390].

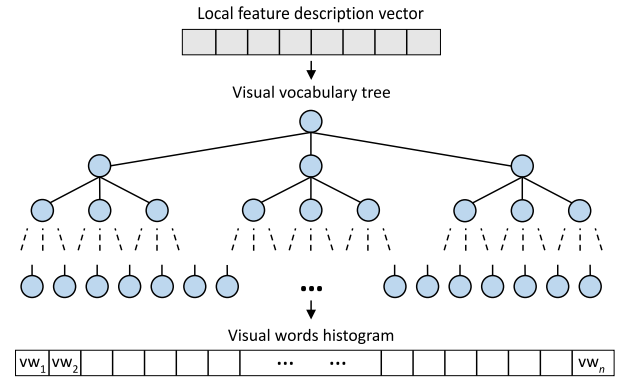

Fig. 15. The structure of a hierarchical visual vocabulary tree used in off-line visual bag of words pipelines [392]. Instead of searching the whole vocabulary to identify the most similar visual word, incoming local feature descriptors traverse the tree significantly reducing the required computations.

Avoiding dimensionality reduction or binary feature vectors, many pipelines were based on GPU-enabled techniques to close loops in real-time with high efficiency [169], [248], [391].

Nister and Stewenius improved the indexing scheme of the off-line visual BoW through a vocabulary tree generated via hierarchical $k$-means clustering [392], as depicted in Fig. 15. This way, faster indexing was achieved, while high performances were preserved [206], [239], [342], [343]. Subsequent works were based on spatial data structures [393] and agglomerative clustering [357]. The inverted multi-index [394] and different tree structures, e.g., $k$-d trees [395], randomized $k$-d forests [396], [397], Chow Liu trees [398], decision trees [399], BK tree [400]. More specifically, data structures, such as pyramid matching [401], [402], were used to detect loop closures when high dimensional image descriptors were adopted [365], [396]. Furthermore, approaches based on the randomized $k$-d forest [76], [79], [90], [110], [119], [373] were shown to perform better than a single $k$-d [70] or a Chow Liu tree [106]. It is worth noting that $k$-d trees are unsuitable when incremental visual vocabularies are selected since they become unbalanced if new descriptors are added after their construction [119]. Yet, this issue is avoided in off-line BoW models since their vocabulary is built a priori, and there is no other time-consuming module regardless of how large the map becomes.

Finally, although impressive outcomes have been achieved by utilizing deep learning, such approaches are yet computationally costly [164]. Increasing the network's size results in more computations and storage consumption at the time of training and testing. However, efforts to reduce their complexity do exist [213]. To bridge the research gap between learned features and their complexity, a CNN architecture employing a small number of layers pre-trained on the scene-centric [403] database reduced the computational and memory costs [153]. Similarly, the authors in [217] compressed the learned features' unnecessary data into a tractable number of bits for robust and efficient place recognition.

## X. Conclusion

Loop closure detection is one of SLAM's most challenging research topics, as it permits consistent map generation and rectification. In this work, we place this problem under a survey focusing on approaches that utilize the camera sensor's input as their primary perception modality. This article revisited the related literature from the topic's early years, where most works incorporated hand-crafted techniques for representing the incoming images, to modern approaches and trends that utilize CNNs to represent the incoming frames. The paper at hand follows a tutorial-based structure describing each of the main parts needed for a visual loop closure detection pipeline to facilitate the newcomers in this area. In addition, a complete listing of the datasets and their features was analytically apposed, while the evaluation metrics were discussed in detail. Closing this survey, the authors wish to note that much effort has been put to produce efficient and robust methods to obtain accurate and consistent maps since the first loop closure detection system. Nonetheless, SLAM and its components remain at the frontline of research, with autonomous robots and driverless cars still evolving. Towards robust map generation and localization, SLAM is able to adopt semantic information regarding the explored environment [404]. For loop closure detection to adapt in such a framework, human-centered semantics of the environment need to be incorporated into its mechanisms. In such a way, long-term autonomy can be facilitated since contextual information allows for a broader hierarchy for organizing visual knowledge. Summarizing the above, future research directions include:

- the development of visual loop closure detection pipelines that operate in dynamic environments which include changing conditions and dynamic scenes;
- performance improvements for severe viewpoint variations;
- improvements along the database's management in order to facilitate long-term mapping.

## Acknowledgment

The authors would like to thank the IEEE TRANSACTIONS ON INTELLIGENT TRANSPORTATION SYSTEMS editorial team, including the editor-in-chief Azim Eskandarian, an associate editor, and all the reviewers, for their prompt, valuable, and constructive feedback, which largely improved the quality of this article.

TSINTOTAS *et al.*: REVISITING PROBLEM IN SLAM: SURVEY ON VISUAL LOOP CLOSURE DETECTION 19951[276] J. Bruce, J. Wawerla, and R. Vaughan, "The SFU mountain dataset: Semi-structured woodland trails under changing environmental conditions," in *Proc. IEEE Int. Conf. Robot. Automat. Workshop*, Seattle, WA, USA, May 2015.

[277] A. Glover, "Day and night, left and right," 2014, doi: 10.5281/zenodo.4590133.

[278] A. J. Glover, W. P. Maddern, M. J. Milford, and G. F. Wyeth, "FAB-MAP + RatSLAM: Appearance-based SLAM for multiple times of day," in *Proc. IEEE Int. Conf. Robot. Automat.*, May 2010, pp. 3507–3512.

[279] W. Maddern, G. Pascoe, C. Linegar, and P. Newman, "1 year, 1000 km: The Oxford RobotCar dataset," *Int. J. Robot. Res.*, vol. 36, no. 1, pp. 3–15, 2017.

[280] G. Ros, L. Sellart, J. Materzynska, D. Vazquez, and A. M. Lopez, "The SYNTHIA dataset: A large collection of synthetic images for semantic segmentation of urban scenes," in *Proc. IEEE Conf. Comput. Vis. Pattern Recognit. (CVPR)*, Jun. 2016, pp. 3234–3243.

[281] M. Ramezani, Y. Wang, M. Camurri, D. Wisth, M. Mattamala, and M. Fallon, "The newer college dataset: Handheld LiDAR, inertial and vision with ground truth," in *Proc. IEEE/RSJ Int. Conf. Intell. Robots Syst. (IROS)*, Oct. 2020, pp. 4353–4360.

[282] D. M. W. Powers, "Evaluation: From precision, recall and F-measure to ROC, informedness, markedness and correlation," 2020, arXiv:2010.16061.

[283] M. Zaffar, A. Khaliq, S. Ehsan, M. Milford, and K. McDonald-Maier, "Levelling the playing field: A comprehensive comparison of visual place recognition approaches under changing conditions," 2019, arXiv:1903.09107.

[284] J. A. Hanley and B. J. McNeil, "The meaning and use of the area under a receiver operating characteristic (ROC) curve," *Radiology*, vol. 143, no. 1, pp. 29–36, 1982.

[285] J. Davis and M. Goadrich, "The relationship between precision-recall and ROC curves," in *Proc. 23rd Int. Conf. Mach. Learn. (ICML)*, 2006, pp. 233–240.

[286] Y. Hou, H. Zhang, and S. Zhou, "Convolutional neural network-based image representation for visual loop closure detection," in *Proc. IEEE Int. Conf. Inf. Automat.*, Aug. 2015, pp. 2238–2245.

[287] B. Ferrarini, M. Waheed, S. Waheed, S. Ehsan, M. J. Milford, and K. D. McDonald-Maier, "Exploring performance bounds of visual place recognition using extended precision," *IEEE Robot. Autom. Lett.*, vol. 5, no. 2, pp. 1688–1695, Apr. 2020.

[288] D. M. Chen et al., "City-scale landmark identification on mobile devices," in *Proc. Conf. Comput. Vis. Pattern Recognit.*, 2011, pp. 737–744.

[289] J. G. Mangelson, D. Dominic, R. M. Eustice, and R. Vasudevan, "Pairwise consistent measurement set maximization for robust multi-robot map merging," in *Proc. IEEE Int. Conf. Robot. Automat. (ICRA)*, May 2018, pp. 2916–2923.

[290] P.-Y. Lajoie, B. Ramtoula, Y. Chang, L. Carlone, and G. Beltrame, "DOOR-SLAM: Distributed, online, and outlier resilient SLAM for robotic teams," *IEEE Robot. Autom. Lett.*, vol. 5, no. 2, pp. 1656–1663, Apr. 2020.

[291] H. Yang, P. Antonante, V. Tzoumas, and L. Carlone, "Graduated non-convexity for robust spatial perception: From non-minimal solvers to global outlier rejection," *IEEE Robot. Autom. Lett.*, vol. 5, no. 2, pp. 1127–1134, Apr. 2020.

[292] Y. Tian, Y. Chang, F. Herrera Arias, C. Nieto-Granda, J. How, and L. Carlone, "Kimera-multi: Robust, distributed, dense metric-semantic SLAM for multi-robot systems," *IEEE Trans. Robot.*, early access, Jan. 20, 2022, doi: 10.1109/TRO.2021.3137751.

[293] H. Lategahn, A. Geiger, and B. Kitt, "Visual SLAM for autonomous ground vehicles," in *Proc. IEEE Int. Conf. Robot. Automat.*, May 2011, pp. 1732–1737.

[294] E. Shechtman and M. Irani, "Matching local self-similarities across images and videos," in *Proc. IEEE Conf. Comput. Vis. Pattern Recognit.*, Jun. 2007, pp. 1–8.

[295] V. Vonikakis, R. Kouskouridas, and A. Gasteratos, "On the evaluation of illumination compensation algorithms," *Multimedia Tools Appl.*, vol. 77, no. 8, pp. 9211–9231, Apr. 2018.

[296] W. Churchill and P. Newman, "Practice makes perfect? Managing and leveraging visual experiences for lifelong navigation," in *Proc. IEEE Int. Conf. Robot. Automat.*, May 2012, pp. 4525–4532.

[297] T. Sattler et al., "Benchmarking 6DOF outdoor visual localization in changing conditions," in *Proc. IEEE/CVF Conf. Comput. Vis. Pattern Recognit.*, Jun. 2018, pp. 8601–8610.

[298] S. Schubert, P. Neubert, and P. Protzel, "Unsupervised learning methods for visual place recognition in discretely and continuously changing environments," in *Proc. IEEE Int. Conf. Robot. Automat. (ICRA)*, May 2020, pp. 4372–4378.

[299] P. Neubert, N. Sünderhauf, and P. Protzel, "Appearance change prediction for long-term navigation across seasons," in *Proc. Eur. Conf. Mobile Robots*, Sep. 2013, pp. 198–203.

[300] A. Ranganathan, S. Matsumoto, and D. Ilstrup, "Towards illumination invariance for visual localization," in *Proc. IEEE Int. Conf. Robot. Automat.*, May 2013, pp. 3791–3798.

[301] S. M. Lowry, M. J. Milford, and G. F. Wyeth, "Transforming morning to afternoon using linear regression techniques," in *Proc. IEEE Int. Conf. Robot. Automat. (ICRA)*, May 2014, pp. 3950–3955.

[302] B. Zhou, A. Lapedriza, J. Xiao, A. Torralba, and A. Oliva, "Learning deep features for scene recognition using places database," in *Proc. Adv. Neural Inf. Process. Syst.*, vol. 27, 2014, pp. 487–495.

[303] B. Bescos, J. M. Fácil, J. Civera, and J. L. Neira, "DynaSLAM: Tracking, mapping, and inpainting in dynamic scenes," *IEEE Robot. Autom. Lett.*, vol. 3, no. 4, pp. 4076–4083, Oct. 2018.

[304] H. Osman, N. Darwish, and A. Bayoumi, "LoopNet: Where to focus? Detecting loop closures in dynamic scenes," *IEEE Robot. Autom. Lett.*, vol. 7, no. 2, pp. 2031–2038, Apr. 2022.

[305] W. Churchill and P. Newman, "Experience-based navigation for long-term localisation," *Int. J. Robot. Res.*, vol. 32, no. 14, pp. 1645–1661, Dec. 2013.

[306] S. M. Lowry, G. F. Wyeth, and M. J. Milford, "Towards training-free appearance-based localization: Probabilistic models for whole-image descriptors," in *Proc. IEEE Int. Conf. Robot. Automat. (ICRA)*, May 2014, pp. 711–717.

[307] Z. Chen, S. Lowry, A. Jacobson, Z. Ge, and M. Milford, "Distance metric learning for feature-agnostic place recognition," in *Proc. IEEE/RSJ Int. Conf. Intell. Robots Syst. (IROS)*, Sep. 2015, pp. 2556–2563.

[308] P. Panphattarasap and A. Calway, "Visual place recognition using landmark distribution descriptors," in *Proc. Asian Conf. Comput. Vis.*, 2016, pp. 487–502.

[309] N. Merrill and G. Huang, "Lightweight unsupervised deep loop closure," in *Proc. Robot., Sci. Syst.*, 2018, pp. 1–10.

[310] N. Carlevaris-Bianco and R. M. Eustice, "Learning visual feature descriptors for dynamic lighting conditions," in *Proc. IEEE/RSJ Int. Conf. Intell. Robots Syst.*, Sep. 2014, pp. 2769–2776.

[311] P. Neubert, N. Sünderhauf, and P. Protzel, "Superpixel-based appearance change prediction for long-term navigation across seasons," *Robot. Auton. Syst.*, vol. 69, pp. 15–27, Jul. 2015.

[312] X. He, R. S. Zemel, and V. Mnih, "Topological map learning from outdoor image sequences," *J. Field Robot.*, vol. 23, nos. 11–12, pp. 1091–1104, Nov. 2006.

[313] C. Linegar, W. Churchill, and P. Newman, "Made to measure: Bespoke landmarks for 24-hour, all-weather localisation with a camera," in *Proc. IEEE Int. Conf. Robot. Automat. (ICRA)*, May 2016, pp. 787–794.

[314] S. Lowry and M. J. Milford, "Supervised and unsupervised linear learning techniques for visual place recognition in changing environments," *IEEE Trans. Robot.*, vol. 32, no. 3, pp. 600–613, Jun. 2016.

[315] J. M. Á. Alvarez and A. M. López, "Road detection based on illuminant invariance," *IEEE Trans. Intell. Transp. Syst.*, vol. 12, no. 1, pp. 184–193, Mar. 2011.

[316] P. Corke, R. Paul, W. Churchill, and P. Newman, "Dealing with shadows: Capturing intrinsic scene appearance for image-based outdoor localisation," in *Proc. IEEE/RSJ Int. Conf. Intell. Robots Syst.*, Nov. 2013, pp. 2085–2092.

[317] M. Shakeri and H. Zhang, "Illumination invariant representation of natural images for visual place recognition," in *Proc. IEEE/RSJ Int. Conf. Intell. Robots Syst. (IROS)*, Oct. 2016, pp. 466–472.

[318] Z. Ying, G. Li, X. Zang, R. Wang, and W. Wang, "A novel shadow-free feature extractor for real-time road detection," in *Proc. 24th ACM Int. Conf. Multimedia*, Oct. 2016, pp. 611–615.

[319] H. Lategahn, J. Beck, B. Kitt, and C. Stiller, "How to learn an illumination robust image feature for place recognition," in *Proc. IEEE Intell. Vehicles Symp.*, Jun. 2013, pp. 285–291.

[320] S. Hausler, A. Jacobson, and M. Milford, "Feature map filtering: Improving visual place recognition with convolutional calibration," in *Proc. Australas. Conf. Robot. Automat.*, 2018, pp. 1–10.

[321] M. Zaffar, S. Ehsan, M. Milford, and K. McDonald-Maier, "CoHOG: A light-weight, compute-efficient, and training-free visual place recognition technique for changing environments," *IEEE Robot. Autom. Lett.*, vol. 5, no. 2, pp. 1835–1842, Apr. 2020.

[322] H. Porav, W. Maddern, and P. Newman, "Adversarial training for adverse conditions: Robust metric localisation using appearance transfer," in *Proc. IEEE Int. Conf. Robot. Automat. (ICRA)*, May 2018, pp. 1011–1018.
Authorized licensed use limited to: University of Thrace (Democritus University of Thrace). Downloaded on November 09,2022 at 11:14:46 UTC from IEEE Xplore. Restrictions apply.

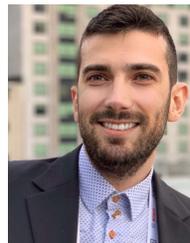

**Konstantinos A. Tsintotas** (Senior Member, IEEE) received the bachelor's degree from the Department of Automation, Technological Education Institute of Chalkida, Psachna, Greece, in 2010, the master's degree in mechatronics from the Department of Electrical Engineering, Technological Education Institute of Western Macedonia, Kila Kozanis, Greece, in 2015, and the Ph.D. degree in robotics from the Department of Production and Management Engineering, Democritus University of Thrace, Xanthi, Greece, in 2021. He is currently a Post-Doctoral Fellow with the Laboratory of Robotics and Automation, Department of Production and Management Engineering, Democritus University of Thrace. His work is supported through several research projects funded by the European Commission and the Greek Government. His research interests include vision-based methods for modern and intelligent mechatronics systems. Details are available at: https://robotics.pme.duth.gr/ktsintotas

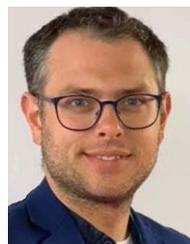

**Loukas Bampis** received the Diploma degree in electrical and computer engineering and the Ph.D. degree in machine vision and embedded systems from the Democritus University of Thrace (DUTh), Greece, in 2013 and 2019, respectively. He is currently an Assistant Professor with the Laboratory of Mechatronics and Systems Automation (MeSA), Department of Electrical and Computer Engineering, DUTh. His work has been supported through several research projects funded by the European Space Agency, the European Commission, and the Greek Government. His research interests include real-time localization and place recognition techniques using hardware accelerators and parallel processing. More details about him are available at: https://robotics.pme.duth.gr/bampis

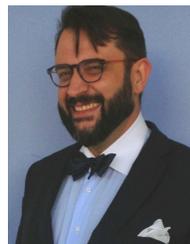

**Antonios Gasteratos** (Senior Member, IEEE) received the M.Eng. and Ph.D. degrees from the Department of Electrical and Computer Engineering, Democritus University of Thrace (DUTh), Greece. He is currently a Professor and the Head of the Department of Production and Management Engineering, DUTh. He is also the Director of the Laboratory of Robotics and Automation (LRA), DUTh, where he is teaching the courses of robotics, automatic control systems, electronics, mechatronics and computer vision. From 1999 to 2000, he was a Visiting Researcher with the Laboratory of Integrated Advanced Robotics (LIRALab), DIST, University of Genoa, Italy. He has served as a reviewer for numerous scientific journals and international conferences. He has published more than 220 papers in books, journals, and conferences. His research interests include mechatronics and robot vision. He is a Fellow Member of IET. He is a Subject Editor of *Electronics Letters* and an Associate Editor of the *International Journal of Optomechatronics* and he has organized/co-organized several international conferences. More details about him are available at: https://robotics.pme.duth.gr/antonis